\documentclass{article}
\usepackage{algorithm} 	
\usepackage{algorithmic}
\usepackage{setspace}
\usepackage{multirow}
 \usepackage{array}
\usepackage{spconf,amsmath,amssymb,epsfig,lipsum,cuted}
\usepackage[colorlinks,
            linkcolor=black,
            anchorcolor=black,
            urlcolor=black,
            citecolor=black]{hyperref}
\usepackage[hyphenbreaks]{breakurl}
\let\OLDthebibliography\thebibliography
\renewcommand\thebibliography[1]{
  \OLDthebibliography{#1}
  \setlength{\parskip}{0pt}
  \setlength{\itemsep}{0pt plus 0.3ex}
}

\begin{document}\sloppy
\def\x{{\mathbf x}}
\def\L{{\cal L}}

\title{Multi-Tier Client Selection for Mobile Federated Learning Networks}
\name{Yulan Gao\textsuperscript{*}, Yansong Zhao\textsuperscript{*}, Han Yu}
\address{Nanyang Technological University,  Singapore\\
\{yulan.gao, billzhao, han.yu\}@ntu.edu.sg
}

\maketitle

\begin{abstract}
Federated learning (FL), which addresses data privacy issues by training models on resource-constrained mobile devices in a distributed manner, has attracted significant research attention.
However, the problem of optimizing FL client selection in mobile federated learning networks (MFLNs), where devices move in and out of each others' coverage and no FL server knows all the data owners, remains open.
To bridge this gap, we propose a first-of-its-kind \underline{Soc}ially-aware
\underline{Fed}erated \underline{C}lient \underline{S}election (SocFedCS) approach to minimize costs and train high-quality FL models.
SocFedCS enriches the candidate FL client pool by enabling data owners to propagate FL task information through their local networks of trust, even as devices are moving into and out of each others' coverage.
Based on Lyapunov optimization, we first transform this time-coupled problem into a step-by-step optimization problem.
Then, we design a method based on alternating minimization and self-adaptive global best harmony search to solve this mixed-integer optimization problem.
Extensive experiments comparing SocFedCS against five state-of-the-art approaches based on four real-world multimedia datasets demonstrate that it achieves 2.06\% higher test accuracy and 12.24\% lower cost on average than the best-performing baseline.
\end{abstract}
\section{Introduction}\label{sec:intro}
Federated learning (FL) is a distributed collaborative machine learning (ML) paradigm that has emerged in the context of growing data privacy concerns. It is built on the fundamental principle of offloading training from a central server to local devices ({\em a.k.a.} clients) \cite{yang2019federated}.
In this way, data owners' sensitive privacy information can be protected as FL avoids the movement of raw data from the devices to the central server.
From an application standpoint, coupled with the prevalence of ubiquitously connected mobile smart devices (e.g., mobile phone, iPad, edge sensors, etc.), the expected upsurge of distributed mobile terminals collaboratively training ML models is materializing \cite{konevcny2016federated}.{\renewcommand{\thefootnote}{}{\fnsymbol{footnote}}\footnotetext{\textsuperscript{*}The authors contributed equally.}}

\begin{figure}[t!]
    \centering
    \includegraphics[width=0.8\columnwidth]{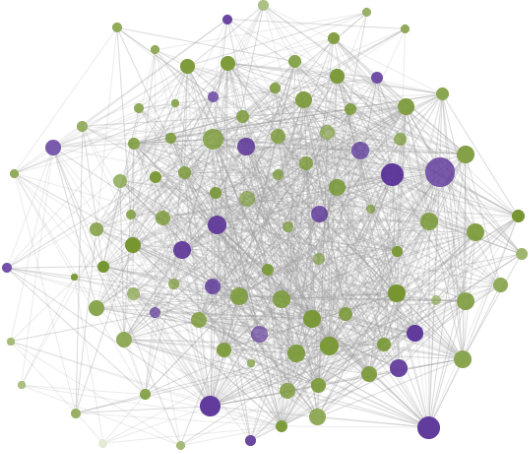}
    \caption{An example MFLN. The size of a node corresponds to the number of other nodes it is connected to. Purple nodes represent FL servers initiating FL tasks. Green nodes represent data owners (i.e., FL clients).}\label{fig:1}
\end{figure}

In mobile federated learning networks (MFLNs)\footnote{\url{https://hacfl.federated-learning.org/network}} as exemplified in Figure \ref{fig:1}, devices (e.g., mobile phones, autonomous vehicles) are highly heterogeneous in terms of their capabilities and availability. This is exacerbated by the fact that mobile devices are moving around and can enter or exit each other's coverage over time. Coupled with concerns that there exist malicious devices seeking to compromise the privacy of others \cite{Lyu2022TNNLS}, optimizing the selection of clients to join FL tasks is an important research challenge.

The social relationships among mobile clients can be established via the mobile social networks (MSNs) connecting them.
Research on leveraging MSNs to enhance FL client selection has emerged in recent years to improve the trustworthiness and reliability of the selected clients, while mitigating misbehaviours \cite{khan2021socially,ottun2022social}.
These investigations, however, are based on the common assumption that the FL server knows all the candidate FL clients.
This assumption is not always realistic, especially in MFLNs in which the inherent conflict between FL client access and wireless broadband shortage exists.
Therefore, considering a more realistic setting in which the FL server does not know all data owners and needs to rely on other nodes to recommend candidate FL clients to it, is a challenging open research question which must be solved to enable practical MFLNs to emerge.

To bridge this important gap, we propose the \underline{Soc}ially-aware \underline{Fed}erated \underline{C}lient \underline{S}election (SocFedCS) approach for trustworthy and cost-effective FL client selection in MFLNs.
It takes into account whether an FL client has established trustworthy social relationships with the FL server, and classifies them into: 1) first-order clients (FCs), and 2) second-order clients (SCs).
The set of FCs are data owners known to the FL server through previous interactions in an MFLN.
In contrast, SCs are data owners who are not directly known to the FL server, but are known to one or more FCs.
The FL server cannot directly access SCs.
The social relationships among nodes in an MFLN can be measured by the frequency of communications.
In addition, FCs and SCs can enter or exit the coverage by a given FL server over time, and thus might not be able to continuously participate in FL training.
SocFedCS aims to dynamically select FL clients for a given FL server from the pool of directly or indirectly known candidate mobile clients to form FL teams under the aforementioned setting.

SocFedCS is, to the best of our knowledge, the first multi-tier client FL selection approach leveraging client recommendation. FCs can recommend trusted SCs to the FL server under SocFedCS to enhance the optimization of client selection decisions.
Specifically, we formulate an infinite-horizon time-average problem to jointly optimize the FL client selection and the local model training to achieve the minimum time-average cost in the worst case.
By leveraging Lyapunov optimization, we transform the complex time-coupled problem into a step-by-step online joint optimization problem.
Specifically, we design a two-stage algorithm which uses a computationally affordable search for obtaining the trustworthy and cost-effective FL client selection, and self-adaptive global best harmony search (SGHS) for optimal local model accuracy.
We evaluate and compare the performance of SocFedCS against five state-of-the-art FL client selection approaches through extensive experiments based on real-world multimedia datasets MNIST, Fashion-MNIST, CIFAR-10, and CIFAR-100.
The results show that SocFedCS achieves $2.06\%$ higher test accuracy and $12.24\%$ lower cost on average than the best-performing baseline.

\section{Related Work}

With regard to FL client selection, a range of sophisticated approaches have been proposed to achieve the goals of improving FL model performance and training efficiency (e.g., FedCS \cite{nishio2019client} and FedGS \cite{li2022data}).
Subsequently, FL client selection strategies for performance trade-off optimization have been studied in \cite{ruan2021valuable}.
An application of the FL client selection in multimedia course recommendation can be found in \cite{qin2022privacy}.
However, they did not take client trustworthiness into account and are not effective in attracting high-quality clients.

A promising way to improve FL client selection is to leverage additional information (e.g., trust among clients).
A pioneering work in \cite{he2019central} revealed the potential benefits of communicating with trusted users for FL through rigorous regret analysis.
Khan {\em et al}. \cite{khan2021socially} investigated the loss function minimization problem for socially-aware-clustering-enabled dispersed FL (DDFL) based on matching theory.
Cluster heads of the DDFL framework are determined based on social ties. They then act as aggregation servers.
The key challenges, opportunities and research roadmap of social-aware FL have been detailed in \cite{ottun2022social}.
To elaborate, FL client selection can be improved by leveraging social connections among clients to identify trusted data owners.
The aforementioned studies are all based on the assumption that the FL server knows all the FL clients, and the FL clients stay within the coverage of the FL server throughout the training process.

Different from the above literature, SocFedCS is designed to
select cost-effective and trustworthy FL clients and schedule them at different rounds, while meeting the long-term requirements of FL training. It supports more realistic settings in which the FL server does not know all the data owners, and enables multi-tier search for suitable clients through a social network \cite{Yu-et-al:2016,Yu-et-al:2017} of client mobile devices which can dynamically enter or exist the coverage area of the FL server.

\section{Preliminaries}

We consider an MFLN consisting of $N$ densely and randomly mobile FL clients within the coverage of the FL server, where the $N$ clients are a union of $M$ FCs and $K$ SCs ($N=M+K$).
The FL server can directly communicate with FCs, but not with SCs.
The index sets of FCs and SCs are denoted as ${\cal M}\triangleq\{1, 2, \ldots, M\}$ and ${\cal K}\triangleq\{1, 2, \ldots, K\}$, respectively.
Without loss of generality, we assume that the number of SCs is no less than FCs (i.e., $K\geq M$).
In order to upload and update trained parameters, we consider an OFDMA protocol to establish a wireless communication for clients.

Discussions on superimposing social networks over mobile FL have been propelled to the forefront of FL research \cite{khan2021socially}.
The driving motivation is that information embeded in social attributes can potentially support client discovery, selection and avoid leakage of learning parameters in MFLNs.
Based on the {\em Community Impact Factor} \cite{khan2021socially}, we characterize the social relationship between FCs and SCs with the level of trust reflected by the average communication frequency between any pair of mobile devices.
We denote social relationships as an $M\times K$ weighted matrix ${\mathbf W}^t=[w_{m,k}^t]$ with each entry $w_{m,k}^t\in[0,1]$ representing the trust between FC $m$ and SC $k$, which is quantified by the normalized communication frequency between $m$ and $k$.
The social trust determines whether an FC would recommend an SC to an FL server or not.
In order to reduce the exposure by an FL server to the risk of unreliable client recommendations, we limit the level of trust transitivity \cite{trusttransitivity2011} to two in this paper (i.e., if a server trusts FC $m$, and $m$ trusts SC $k$, then the server can trust $k$; but no further).

FL training minimizes the loss function through communications between the server and the clients.
Each client iteratively computes the local model until a local accuracy $0\leq \theta\leq 1$ is achieved, and then uploads it to the FL server.
The collected local models are aggregated to form a global model $\pmb\omega$.
When a specific global model accuracy $0\leq \epsilon\leq 1$ is reached, the FL training process is terminated.
As mentioned in \cite{tran2019federated}, for the convex loss function, the local iterations is general upper bounded by ${\cal O}(\log(1/\theta))=\eta\log(1/\theta),$ which is suitable for gradient or stochastic descent iterative algorithms.
In addition, the number of local iterations can be normalized as $\log(1/\theta).$
Here, $\theta$ and $\epsilon$ are the differences of the model gradients between two successive iterations of training at the local level and at the global level, respectively.{\footnote{$||\triangledown{\cal L}({\pmb\omega}^t)||\leq \epsilon||\triangledown{\cal L}({\pmb\omega}^{t-1})||$ and $||\triangledown{\cal L}_m({\pmb\omega}^t)||\leq \theta||\triangledown{\cal L}_m({\pmb\omega}^{t-1})||$, where ${\cal L}(\pmb\omega)$ and ${\cal L}_m(\pmb\omega)$ are the loss function of global and local model, respectively.}}

\section{The Proposed Approach}
In this section, we describe how SocFedCS leverages social relationships among MFLN devices to optimize FL client selection.
To facilitate the introduction of SocFedCS, the following entities are modeled from the perspective of FCs because SCs are recommended by FCs known to the FL server.
Thus, the union of potential clients in the view of an FC $m$ is $\bar{\cal N}_m={\cal N}_m\cup\{m\}$, where ${\cal N}_m=\{k\in{\cal K}~| w_{m,k}>0\}$ is the set of SCs trusted by $m$.
The FL server can directly select a known FC or an SC recommended by a known FC.

In the communication phase, the data rate of each FL client $i\in\bar{\cal N}_m, m\in{\cal M}$ under SocFedCS can be written as:
\begin{align}\label{eq:2}
R_{m,i}^t=\alpha_{m,i}^tB\log[1+{h_{i}^tp_i}/{(N_0B)}],
\end{align}
where $\alpha_{m,i}^t\in\{0, 1\}$ is the FL client selection indicator in the view of FC $m$ in the $t\text{-th}$ round, i.e., $\alpha_{m,i}^t=1$ if any $i\in\bar{\cal N}_m$ is chosen by the server, $\alpha_{m,i}^t=0$ otherwise. $B$ represents the transmission bandwidth, $N_0$ is the spectrum density of the white Gaussian noise. $p_i$ represents the transmit power budget of client $i$. $h_{i}^t$ denotes the channel power gain between client $i$ and the FL server.
We denote the data size of the local model at client $i$ as $C_i$. Thus, the time of uploading them can be estimated as $T_{m,i}^{\text{com}, t}=C_i/R_{m,i}^t$.
Correspondingly, the energy consumption for uploading local model parameters is $E_{m,i}^{\text{com},t}=\alpha_{m,i}^tp_iT_{m,i}^{\text{com},t}.$

In the computation phase, the total number of CPU cycles required to process the training $D_i$ data samples is $D_iQ_i$, where $Q_i$ is the number of CPU cycles that is required to process one data sample.
Let $f_i$ be the CPU frequency of client $i$, $\rho_if_i^{\zeta}$ denotes the computational power of $i$, where $\rho_i$ is a constant that depends on the average switched capacitance and the average activity factor, and  $\zeta (\zeta\geq 2)$ is a constant.
Then, under SocFedCS, the computation time for one local iteration at client $i$ is $T_{m,i}^{\text{cmp},t}={\alpha_{m,i}^tD_iQ_i}/f_i$.
The corresponding energy consumption for one local iteration at client $i$ is $E_{m,i}^{\text{cmp}, t}=\rho_iD_iQ_i(\alpha_{m,i}^t)^{\zeta-1}f_i^{\zeta-1}$.
The time cost, $T_{m,i}^t$, and energy consumption, $E_{m,i}^t$, at client $i\in\bar{\cal N}_m$ in the $t\text{-th}$ round can be respectively formulated as:
\begin{align}\label{eq:4}
T_{m,i}^t&={1}/{(1-\theta^t)}[\log(1/\theta^t)T_{m,i}^{\text{cmp},t}+T_{m,i}^{\text{com},t}],\\
E_{m,i}^t&={1}/{(1-\theta^t)}[\log(1/\theta^t)E_{m,i}^{\text{cmp},t}+E_{m,i}^{\text{com},t}].
\end{align}

{\bf\em Constraints and Assumptions. }
{\em (1) Long-term Goal of FL training:}
For a given MFLN, the long-term goal of FL training shall be considered as FCs may occasionally be unable to perform FL tasks due to resource constraints and unavailability.
To this end, we introduce a long-term constraint in Eq. \eqref{eq:5} to ensure that the target participation rate for an FC $m$ shall not be lower than $\Delta\in[0,1]$.
Without loss of generality, we set $\Delta=L/N$ where $L$ is the maximum number of clients allowed to join FL training in each round.
\begin{align}\label{eq:5}
\varlimsup_{R\rightarrow\infty}\frac{1}{R}\sum\nolimits_{t=1}^R{\mathbb E}\Big\{\sum\nolimits_{i\in{\bar{\cal N}_m}}\alpha_{m,i}^t\Big\}\geq \Delta, \forall m\in{\cal M}.
\end{align}

{\em (2) The Duration of a Local Iteration:}
To alleviate the straggler issue, for each FL client $i\in\bar{\cal N}_m, m\in{\cal M}$, the training time for one local iteration shall be upper bounded by $T_{\max}^{\text{cmp}}$, which can be formulated as:
\begin{align}\label{eq:6}
\alpha_{m,i}^t{D_i^tQ_i}/{f_i}\leq T_{\max}^{\text{cmp}}, \forall i\in\bar{\cal N}_m, m\in{\cal M}.
\end{align}
We employ the weighted sum method to deal with the trade-off between energy consumption and delay using parameters $\lambda_i^t$ and $\lambda_i^e$ with $0\leq \lambda_i^t, \lambda_i^e\leq 1, \lambda_i^t+\lambda_i^e=1$.
Thus, the Weighted Sum of Energy consumption and Time cost (WSET) is expressed as $\lambda_i^tT_{m,i}^t+\lambda_i^eE_{m,i}^t$.
Therefore, the total cost of each $i\in\bar{\cal N}_m, m\in{\cal M}$ can be modeled as:
\begin{align}\label{eq:12}
{\cal G}_{m,i}^t=\begin{cases}
 \lambda_i^tT_{m,i}^t+\lambda_{i}^eE_{m,i}^t+\sigma C_0, &\text{if~} i\neq m,\\
 \lambda_i^tT_{m,i}^t+\lambda_{i}^eE_{m,i}^t, &\text{otherwise},
\end{cases}
\end{align}
where $\sigma C_0$ represents the cost of an SC being successfully recommended by an FC. $\sigma>0$ is a control parameter that enables system administrators to indicate their relative preferences between WSET and the cost of recommendation.
It is reasonable to add an {\em additional  cost} when an SC is selected since SCs are not directly known to an FL server.
In addition, for simplicity, $C_0>0$ can be considered as constant since FCs have trusted SCs within their local networks.

To summarize,  SocFedCS is designed to solve the following optimization problem:
\begin{align}
\min_{{\pmb\alpha}^t, \theta^t}&\varlimsup_{R\rightarrow \infty} \frac{1}{R}\sum\nolimits_{t=1}^R\max_{i\in\bar{\cal N}_m, m\in{\cal M}} {\cal G}_{m,i}^t \label{eq:7}\\
\text{s.t.~}&~\alpha_{m,i}^t\in\{0,1 \}, \forall i\in\bar{\cal N}_m, m\in{\cal M},\tag{\ref{eq:7}a}\\
&~\sum\nolimits_{m=1}^M\alpha_{m,i}^t\leq 1, \forall i\in\bar{\cal N}_m,\tag{\ref{eq:7}b}
\end{align}
\begin{align}
&~\sum\nolimits_{i\in\bar{\cal N}_m}\alpha_{m,i}^t\leq 1, \forall m\in{\cal M},\tag{\ref{eq:7}c}\\
&~~\text{Eq. }\eqref{eq:5} \text{~and~Eq. } \eqref{eq:6},\tag{\ref{eq:7}d}
\end{align}
where $\pmb\alpha^t=\{\alpha_{m,i}^t\}_{i\in\bar{\cal N}_m,m\in{\cal M}}$ represents the set of client selection policies.

The goal of Eq. \eqref{eq:7} is to minimize clients' long-term cost (WSET and {\em cost} of recommendation) subject to the long-term requirements of FL training in Eq. \eqref{eq:5} (which tolerates short-term violation), as well as a hard constraint in Eq.  \eqref{eq:6} which cannot be compromised.
The time-coupled objective function and Constraint \eqref{eq:5} are  difficult for offline solutions to deal with.
In addition, client selection is performed before the real training process.
Subsequently, for an alternative suboptimal problem to be solved, we elaborate on our transformation of the offline problem into a step-by-step online scheduling problem which can be effectively solved by Lyapunov optimization \cite{yu2020sustainable}.

To transform Eq. \eqref{eq:7} into an online optimization problem,
we introduce a virtual queue $z_m^t$ for an FC $m$ to re-write Eq. \eqref{eq:5}.
The queuing dynamics of $z_m$ is expressed as $z_m^{t+1}=\max\left[0,~ z_m^t + \Delta1_{[\sum_{i\in\bar{\cal N}_m}\alpha_{m,i}^t=0]}-\sum_{i\in\bar{\cal N}_m}\alpha_{m,i}^t\right]$.
$1_{[\text{condition}]}$ is an indicator function. Its value is $1$ if [condition] is true; otherwise, its value is $0$.
Inspired by \cite{yu2020sustainable},  Eq. \eqref{eq:5} holds if all the virtual queues remain mean rate stable throughout the FL process.
Thus, we use Lyapunov optimization to limit the growth of each queue $z_m^t$.
Consider a quadratic Lyapunov function $L(\pmb\Theta^t)\triangleq\frac{1}{2}\sum_{m\in{\cal M}}(z_m^t)^{2},$ where $\pmb\Theta^t\triangleq\{z_m^t\}_{m\in{\cal M}}.$
The drift of the Lyapunov function $\bigtriangleup L({\pmb\Theta}^t)={\mathbb E}[L({\pmb\Theta}^{t+1})-L({\pmb\Theta}^{t})|{\pmb\Theta}^t]$ satisfies:
\begin{align}\label{eq:13}
\bigtriangleup L(\pmb\Theta^t)\leq \Gamma+\hspace{-0.5em}\sum\nolimits_{m\in{\cal M}}z_m^t{\mathbb E}\Big[\Delta-\hspace{-0.5em}\sum\nolimits_{i\in\bar{\cal N}_m}\alpha_{m,i}^t|\pmb\Theta^t\Big],
\end{align}
where $\Gamma={M(1+\Delta^2)}/{2}$ is a constant.
Thus, our goal is to minimize the combination of the right-hand-side of Eq. \eqref{eq:13} and a penalty term $V{\cal G}_{m,i}^t$, where $V\geq 0$ is a weight for balancing the trade-off between the goals of minimization and satisfying Eq. \eqref{eq:5}.

For brevity, we omit the constant (i.e., $\Gamma$) and transform Eq. \eqref{eq:7} into the following problem:
\begin{align}
&\min_{{\pmb\alpha}^t, \theta^t}\Big\{\max_{i\in\bar{\cal N}_m, m\in{\cal M}}V{\cal G}_{m,i}^t+\hspace{-0.5em}\sum_{m\in{\cal M}}z_m^t\Big[\Delta-\hspace{-0.5em}\sum_{i\in\bar{\cal N}_m}\alpha_{m,i}^t\Big]
\Big\}\label{eq:8}\\
&~~\text{s.t.~Eq. }\eqref{eq:6}\text{~and~Eq. }(\ref{eq:7}a)-(\ref{eq:7}c). \tag{\ref{eq:8}a}
\end{align}
It can be observed that the objective function Eq. \eqref{eq:8} is non-convex and Constraints (\ref{eq:7}a) and (\ref{eq:7}b) imply that Eq. \eqref{eq:8} is an integer optimization problem.
In general, there is no efficient and standard method to solve such mixed-integer problems.
Using the alternating optimization techniques, we separately and iteratively solve $\pmb\alpha^t$ and $\theta^t$.
In particular, we first solve for optimal FL client selection $\pmb\alpha^t$ given a fixed $\theta^t$, and then derive the optimal local model accuracy $\theta^{t*}$ when $\pmb\alpha^t$ is fixed.
Moreover, due to the unique client selection constraints Eq.~(\ref{eq:7}a)--Eq.~(\ref{eq:7}c), the number of possible client selection strategies is reduced from $2^{M(K+1)}$ to $(K+1)^M$.

Here, given $\theta^t$, we propose a computationally affordable approach according to the characteristics of Constraint \eqref{eq:6} to obtain the feasible set of selected clients for Eq. \eqref{eq:8}.
With this feasible selection set, upon introducing $F_{m,i}(\theta^t)=V{\cal G}_{m,i}^t+\sum_{m\in{\cal M}}[\Delta-\sum_{i\in\bar{\cal N}_m}\alpha_{m,i}^t]$,
each FC $m$ simply ranks candidate SCs ($i\in\bar{\cal N}_m$) in ascending order of their $F_{m,i}(\theta^t)$ values, and recommends the one with the lowest $F_{m,i}(\theta^t)$ to the FL server.
The FL server then sorts the received candidate SCs in ascending order of their $F_{m,i}(\theta^t)$ values.
For each round, the FL server selects the top-$L$ clients with the lowest $F_{m,i}(\theta^t)$ values for FL training.
Then, the FL server updates the virtual queues associated with all FCs.

Based on the aforementioned optimal selection policy $\pmb\alpha^t$, we set the function ${\cal F}(\theta^t)=\max_{i\in\bar{\cal N}_m, m\in{\cal M}} F_{m,i}(\theta^t)$.
Upon introducing function ${\cal F}(\theta^t)$, the optimal local model accuracy can be deduced by the following unconstrained problem:
\begin{align}\label{eq:10}
\min_{0\leq \theta^t\leq 1} {\cal F}(\theta^t).
\end{align}
Eq. \eqref{eq:10} can be solved by following the SGHS algorithm \cite{mahdavi2007improved}.
It is described by a five-tuple $\langle\text{HMS}, \text{HMCR}, \text{PAR}, \text{BW}, \text{NI}\rangle$ consisting of the following components: Harmony Memory Size (HMS), Harmony Memory Consideration Rate (HMCR), Pitch Adjustment Rate (PAR), Distance BandWidth (BW) and the number of Improvisations (NI).
By introducing these adaptive parameters, SGHS can achieve improved performance on continuously optimizing Eq.  \eqref{eq:10}.
\setcounter{table}{0}
\begin{table*}[!t]
  \centering
  \caption{Performance comparison under Scenario 1 (S1) and Scenario 2 (S2).}
  \label{table:1}
  \setlength{\tabcolsep}{1mm}{
\begin{tabular}{|c|c|c|c|c|c|c|c|c|c|c|c|c|}
\hline
\multirow{3}{*}{Method}&\multicolumn{3}{c|}{MNIST}&\multicolumn{3}{c|}{CIFAR-10}&\multicolumn{3}{c|}{Fashion-MNIST}&\multicolumn{3}{c|}{CIFAR-100} \\
\cline{2-13}
{~}& {~}& S1 & S2 &  & S1 & S2& &S1 &S2 & & S1 &S2 \\
\cline{3-4}  \cline{6-7} \cline{9-10} \cline{12-13}
& Cost & Acc (\%) & Acc (\%) & Cost & Acc (\%) & Acc (\%)& Cost & Acc (\%) & Acc (\%)& Cost & Acc (\%) & Acc (\%) \\
\hline
Random & 3.981 & 85.21 & 67.48 & 4.247 & 58.93 & 49.33 &3.996&67.23&60.52 &4.276 &47.82& 38.27\\
Greedy & 2.976 & 94.15 & 75.17 & 3.257 & 64.06 &57.70 & 2.966 & 83.09 &74.98 &3.265 &50.77 &44.25\\
PowCS & 3.987 & 96.25 & 82.63 & 4.248 & 65.71 & 57.13 &3.988 & 95.51 & 82.14 & 4.289 & 55.27 & 49.27 \\
FedCS & 4.997 &  {\bf 97.98} & 85.42 & 4.299
& 65.33 & 57.07 & 5.017 & 94.17 & 85.01 & 5.349 & 57.21 & 48.37\\
Oort & 3.568 & 96.33 & 83.73 & 3.848 & 65.96 & 56.34 & 3.670 & 95.12 & 81.76 & 3.975 & 55.22 &  49.33\\
\hline
SocFedCS & {\bf 2.598} & 97.70&  {\bf 87.09} & {\bf 2.901} & {\bf 68.49} & {\bf 58.51} & {\bf 2.490} & {\bf 97.56} & {\bf 85.66} & {\bf 2.950} & {\bf 60.17} & {\bf 51.83}\\
\hline
   \end{tabular}}
\end{table*}

\section{Experimental Evaluation}
In this section, we conduct extensive experiments to evaluate SocFedCS based on the Erdos-Renyi random network \cite{crucitti2004model} and four multimedia datasets: MNIST, Fashion-MNIST, CIFAR-10, and CIFAR-100.

\subsection{Experiment Settings}
To ensure that the complexity of the simulations is tractable while considering a significantly loaded system, we design a circular area with size $\pi(100\times 100)\text{~m}^2$ centered around an FL server containing $M=40$ randomly distributed FCs and $K=80$ SCs.
To model a realistic mobile environment, the Gauss-Markov mobility model \cite{camp2002survey} is used to simulate the movements of mobile devices.
The Erdos-Renyi random graph is used to construct the social relationships among FCs and SCs with a connection probability of $p=0.7$.
The trust level ${\mathbf W}^t$ can be further established through online social networks, the entries $w_{m,k}^t$ of which quantify the average communication rate between pairs of nodes.

We consider an MFLN environment with carrier frequency $2.3\text{~GHz}$ and bandwidth $0.2\text{~MHz}$.
The channel realizations are generated according to the 3GPP propagation environment.
Throughout the simulations, candidate clients are assumed to have the same computation time constraint in one local iteration, $\text{T}_{\max}^{\text{cmp}}=0.1\text{~s}$.
The transmit power budget $p_n$ of each client $n\in{\cal N}$ is randomly assigned from the set $\{0.1, 0.2, 0.3, 0.4, 0.5\}\text{~Watt}$.
The CPU frequency $f_n$ of each client $n$ is uniformly selected from the set $\{ 2\times 10^7, 3\times 10^7, \ldots, 2\times 10^8\}\text{~cycles/byte}.$

For mentioned datasets, the training set is distributed to $N=120$ clients (i.e., FCs and SCs).
The availability of each client follows the same Bernoulli distribution with parameter $0.6$.
For each round, we select $L=14$ clients from the population. We investigated two experimental scenarios:
{\bf Scenario 1:}  the data distribution of clients is IID. To simulate different quality levels of local data, we generate different percentages of noisy labels to clients based on the trust levels between FCs and SCs. Specifically, each FC $m$ owns $100(1-w_{m,:}/w)\%$ noisy data, where $w_{m,:}=\sum_{k\in{\cal K}}w_{m,k}$ and $w=\sum_{m\in{\cal M}, k\in{\cal K}}w_{m,k}$.
Likewise, each SC $k$ owns $100(1-w_{:,k}/w)\%$ noisy data with $w_{:,k}=\sum_{m\in{\cal M}}w_{m,k}$.
{\bf Scenario 2:} following the data quality settings in Scenario 1, we consider the non-IID data distributions with a heterogeneity level of $30\%$.

\subsection{Comparison Approaches}
We compare the performance of SocFedCS with the following five approaches in terms of the time-average cost (Eq. \eqref{eq:7}) and the global model test accuracy:
\begin{enumerate}
\setlength{\itemsep}{1pt}
\setlength{\parskip}{1pt}
\setlength{\parsep}{1pt}
\item{\bf Random}: FL server randomly selects $L$ available FCs.
\item {\bf Greedy}: FL server selects the top $L$ available FCs with the lowest costs in each round.
\item{\bf PowCS} \cite{cho2020client}: Under this approach, the server first samples from a fixed candidate set ${\cal M}_0$ ($L\leq |{\cal M}_0|\leq M$) FCs. Then, the server sends the current global model $\pmb\omega^t$ to the FCs in ${\cal M}_0$. These FCs compute and send back to the server their local loss.{\footnote{The local loss function is ${\cal L}_m(\pmb\omega)=\frac{1}{D_m}\sum_{l=1}^{D_m}f_l({\boldsymbol x}_l, {\boldsymbol y}_l, {\pmb\omega})=\frac{1}{D_m}\sum_{l=1}^{D_m}\frac{1}{2}||{\boldsymbol y}_l-{\pmb\omega}^T{\boldsymbol x}_l||^2$, where $\{{\boldsymbol x}_l, {\boldsymbol y}_l\}_{l=1}^{D_m}$ represents the input-output pair of each data sample.}}
Finally, from the candidate set ${\cal M}_0$, the server selects the $L$ FCs with the highest local loss.
\item{\bf FedCS} \cite{nishio2019client}: It allows the FL server to aggregate as many available FC local model updates as possible within a fixed deadline.
We set the deadline to $2\text{~s}$, which makes the number of selected clients appropriate relative to $L$.
\item{\bf Oort} \cite{lai2021oort}: Based on the guidelines outlined in  \cite{lai2021oort}, the FL server possesses the capacity to select the top $L$ FCs who have made the most valuable contributions.
\end{enumerate}

The performance comparison is based on the premise that the aforementioned approaches only perform FL client selection in the first-order client tier.
To assess the effectiveness of SocFedCS, we set up an environment with PyTorch.
We train two different convolutional neural network (CNN) models for MNIST, Fashion-MNIST, CIFAR-10, and CIFAR-100, respectively.
All mentioned datasets are divided among the FL clients following Scenario 1 and Scenario 2.

\subsection{Results and Discussion}
In this section,  we analyze the performance of the comparison approaches in terms of time-average cost and test accuracy. The results are shown in Table \ref{table:1}.

In terms of time-average cost, it can be observed that SocFedCS significantly outperforms all the baselines.
FedCS performs the worst due to its aggressive strategy for selecting as many clients as possible, which can lead to significant increases in costs.
SocFedCS can reduce cost by $12.24\%$ on average compared with the best performing baseline - Greedy - which can only perform FL client selection on the FC tier.
This result demonstrates that, despite the additional cost of cross-tier client selection, the SocFedCS approach enables multi-tier search for suitable clients with minimal cost.

According to Eq. \eqref{eq:12}, the FL server's aggressive attitude towards cost reduction can lead to a higher local test accuracy (i.e., $\theta$), thereby leading to lower global test accuracy.
As expected, the model performance under Greedy, which only performs FL client selection on the FC tier, is poor.
SocFedCS can consistently produce high-quality FL models, while reducing the cost.
Table \ref{table:1} shows that SocFedCS achieves the highest test accuracy or close to the highest test accuracy under all experiment settings. It achieves $2.06\%$ higher average test accuracy than the best performing baseline, FedCS.

\section{Conclusions and Future Work}
In this paper, we proposed the SocFedCS to optimize multi-tier FL client selection in MFLNs, in which FL servers do not know all candidate FL clients.
We design a virtual queue and leverage Lyapunov optimization to tackle the long-term constraints and objectives that are time-coupled.
A computationally affordable iterative algorithm exploiting alternating minimization and SGHS is then proposed to solve the mixed-integer optimization problem efficiently.
In light of the results of simulations and experiments, we found that SocFedCS achieved $12.24\%$ lower cost and $2.06\%$ higher test accuracy on average than the best performing baselines. To the best of our knowledge, it is the first multi-tier client selection approach designed for MFLNs.

In subsequent research, we will extend SocFedCS to support longer chains of trust transtivity beyond the current two steps so that more candidate FL clients can be indirectly leveraged in complex MFLNs.

\section{Acknowledgement}
This research is supported, in part, by the National Research Foundation Singapore and DSO National Laboratories under the AI Singapore Programme (AISG Award No: AISG2-RP-2020-019); the RIE 2020 Advanced Manufacturing and Engineering (AME) Programmatic Fund (No. A20G8b0102), Singapore; the Joint NTU-WeBank Research Centre on Fintech (NWJ-2020-008); the Nanyang Assistant Professorship (NAP); Future Communications Research \& Development Programme (FCP-NTU-RG-2021-014); and the Joint SDU-NTU Centre for Artificial Intelligence Research (C-FAIR).


\end{document}